\documentclass[10pt,twocolumn,letterpaper]{article}

\usepackage{wacv}
\usepackage{times}
\usepackage{epsfig}
\usepackage{graphicx}
\usepackage{amsmath}
\usepackage{amssymb}
\usepackage{booktabs}
\usepackage{multirow}
\usepackage[table,xcdraw]{xcolor}
\usepackage[normalem]{ulem}
\useunder{\uline}{\ul}{}

\usepackage[accsupp]{axessibility}
\def\wacvPaperID{145} % Enter the WACV Paper ID here

\wacvalgorithmstrack   % Uncomment this line if you are submitting to the Algorithms Track.
\wacvfinalcopy % *** Uncomment this line for the final submission

\ifwacvfinal
\usepackage[breaklinks=true,bookmarks=false]{hyperref}
\else
\usepackage[pagebackref=true,breaklinks=true,colorlinks,bookmarks=false]{hyperref}
\fi

\begin{document}

%%%%%%%%% TITLE
\title{Multi-view 3D Object Reconstruction \\ and Uncertainty Modelling with Neural Shape Prior
}

\author{Ziwei Liao and Steven L. Waslander\\
Robotics Institute, Institute for Aerospace Study\\
University of Toronto\\
{\tt\small ziwei.liao@mail.utoronto.ca, \tt\small steven.waslander@utoronto.ca}
}

\maketitle
\thispagestyle{empty}

%%%%%%%%% ABSTRACT
\begin{abstract}
3D object reconstruction is important for semantic scene understanding. It is challenging to reconstruct detailed 3D shapes from monocular images directly due to a lack of depth information, occlusion and noise. Most current methods generate deterministic object models without any awareness of the uncertainty of the reconstruction. We tackle this problem by leveraging a neural object representation which learns an object shape distribution from large dataset of 3d object models and maps it into a latent space. We propose a method to model uncertainty as part of the representation and define an uncertainty-aware encoder which generates latent codes with uncertainty directly from individual input images. Further, we propose a method to propagate the uncertainty in the latent code to SDF values and generate a 3d object mesh with local uncertainty for each mesh component. Finally, we propose an incremental fusion method under a Bayesian framework to fuse the latent codes from multi-view observations. We evaluate the system in both synthetic and real datasets to demonstrate the effectiveness of uncertainty-based fusion to improve 3D object reconstruction accuracy.

\end{abstract}

\begin{figure}[t]
\centering

\includegraphics[width=0.47\textwidth]{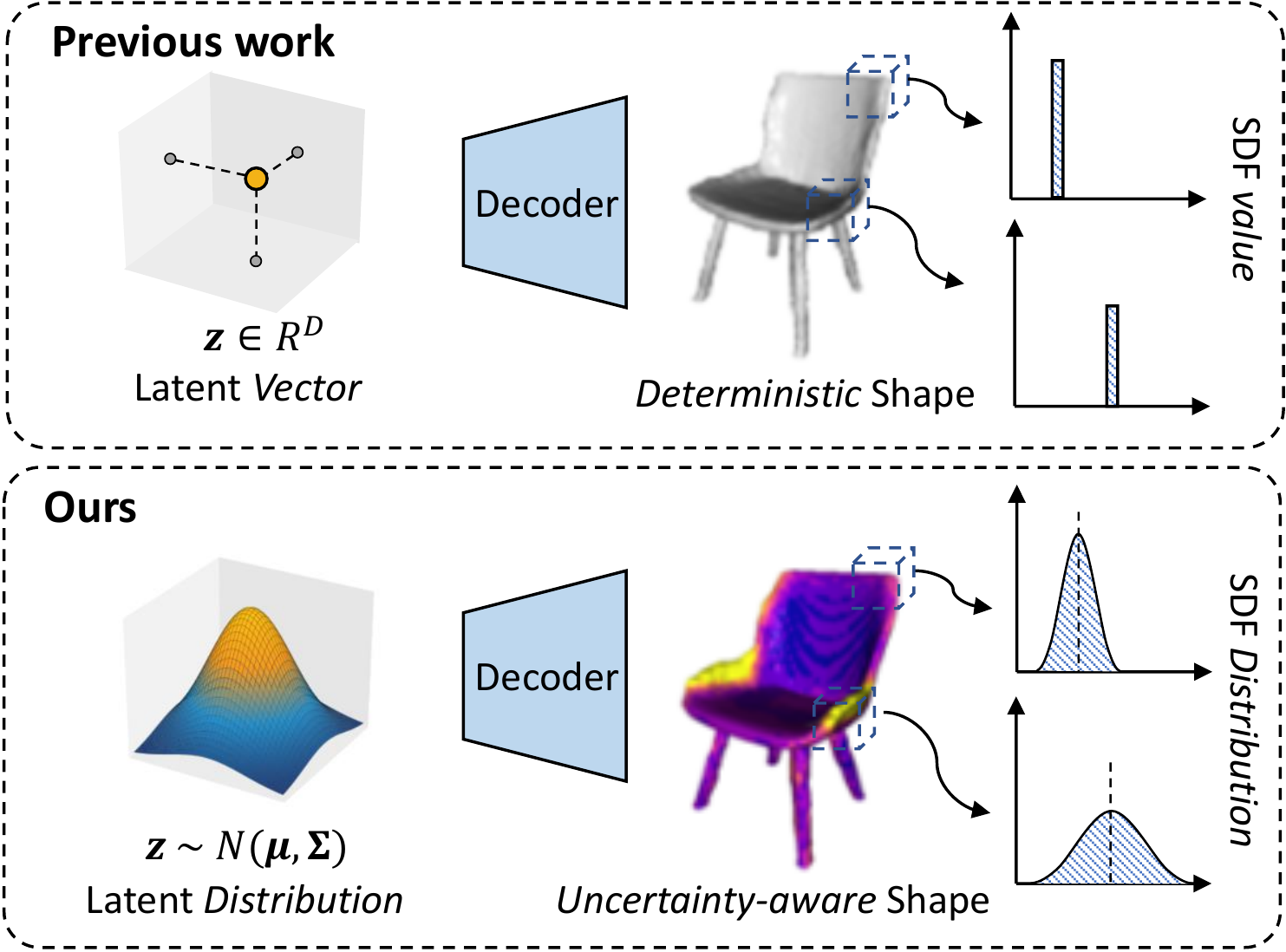}

\caption{The proposed uncertainty-aware 3D object representation. While the previous work decodes a deterministic shape from a latent vector, Ours decodes a shape with uncertainty from a latent distribution. We can reconstruct 3D objects with uncertainty attached to each part from monocular or multi-view images.}
\label{fig:cover}
\end{figure}

%%%%%%%%% BODY TEXT
\section{Introduction}

Identifying and modelling 3D objects in the scene is an important step towards semantic scene understanding~\cite{naseer2018indoor}. Accurate object representations are key elements for down stream tasks such as object detection, segmentation, tracking, manipulation and dynamic change detection. However, reconstructing detailed 3D object shapes from limited image data remains challenging~\cite{choy20163d,fan2017point,wu2016learning,wu2017marrnet,groueix2018papier,sun2018pix3d,gkioxari2019mesh,runz2020frodo,liu2021voxel}. Man-made objects have highly variable shapes. Monocular observations can be degraded by issues such as occlusion, noise, truncation, lack of depth measurements, which makes the reconstruction task a ill-posed problem. Some form of prior knowledge of object shape is needed, and has been shown to significantly improve reconstruction performance from single \cite{liu2021voxel} and multiple viewpoints \cite{runz2020frodo}.

This paper aims to propose a robust 3D object reconstruction method based on monocular images. To address the problems mentioned above, we concentrate on efficiently and robustly combining object priors  through fusion of multi-view observations. In particular, we propose an uncertainty-aware fusion framework that refines 3D object representations as new viewpoints are observed, and exploits large object shape datasets to maximize prior knowledge of object geometries. 

Recently, neural implicit representations \cite{mildenhall2020nerf} have presented remarkable achievements in multiple areas, including object reconstruction~\cite{runz2020frodo}, object SLAM~\cite{wang2021dsp,sucar2020nodeslam}, and scene reconstruction~\cite{sucar2021imap,zhu2022nice,zhu2023nicer}. Neural object representations, e.g., DeepSDF \cite{park2019deepsdf}, OccupancyNet \cite{mescheder2019occupancy}, show the ability for interpolation, partial completion of 3d shapes and reconstruction from point cloud or monocular images \cite{mescheder2019occupancy,runz2020frodo}. The representation can be trained on object shape datasets to learn a prior distribution of a specific class of objects. 

We take a step further to couple uncertainty into the neural object representation as in Fig.~\ref{fig:cover}. Uncertainty modelling is highly critical for AI safety and robotics. For example, autonomous driving vehicles need to make safe and reliable decisions based on incomplete or noisy data. The uncertainty accumulated in the perception system can be propagated to the down stream tasks to help applications like localization, tracking, motion planning, and make system maximize the usage of the multi-view observations with robustness to corrupted observations. We discuss more uncertainty applications in the Supplementary Materials.

Uncertainty modelling in deep learning is a widely studied area. Methods including Bayesian Neural Networks~\cite{mackay1992practical}, sampling approaches (e.g., MC Dropout~\cite{gal2016uncertainty}, Ensembles~\cite{lakshminarayanan2017simple}) and direct methods~\cite{kendall2017uncertainties}, have been proposed and used in real applications such as object detection \cite{feng2021review} and semantic segmentation \cite{huang2018efficient}. As far as we know, this paper is the first time to estimate uncertainty for neural object representation from monocular images. 

To couple uncertainty, we propose a framework that can propagate uncertainty from image space, to latent space, and finally to 3D object shape, as in Fig. \ref{fig-framework}. Specifically, we propose a way to teach the encoder to produce a code uncertainty that leads to the right model uncertainties from single images. Then, we propose a method to propagate the uncertainty through the decoder to the SDF and onto the mesh. We design a two-stage training strategy following the previous work~\cite{runz2020frodo}. First, we train the decoder to learn a latent space. Then, holding the decoder fixed, we force the encoder to output the correct code uncertainty. This strategy makes the encoder and decoder loosely coupled, and stores the uncertainty in the latent space, which can in theory generalize to different types of decoders.
% summarize contributions here
We summarize our contributions below:
\begin{itemize}
    \item We propose a 3d object modelling approach that relies on an implicit neural representation and provides both a 3D object reconstruction and an uncertainty measure for each object.
    \item We propose an image encoder with direct uncertainty modelling to estimate latent codes with uncertainty from a single image.
    \item We propose an incremental fusion method that relies on Bayesian inference to fuse multi-view observations in the latent space to improve reconstruction accuracy and reduce spatial uncertainty.
    \item We evaluate the system in both synthetic and real datasets, demonstrating the benefit of fusing object models produced from different views through Bayesian inference on the encoded representation.
\end{itemize}

\section{Related Work}

\subsection{3D Object Representations and Reconstruction}

Common 3D object representations include meshes~\cite{gkioxari2019mesh}, voxels~\cite{liu2021voxel}, octrees~\cite{zeng2013octree}, TSDFs~\cite{park2019deepsdf} and point clouds~\cite{fan2017point}, which are all flexible representations but require heavy storage and computation. Each element in the representation is discrete and independent, thus it remains difficult to reconstruct detailed shape from partial observations. There are also compact representations using geometric primitives such as cuboids and quadrics \cite{yang2019cubeslam,nicholson2018quadricslam,liao2022so},  which are significantly more computation efficient but only provide limited information for localization and insufficient information for collision detection and manipulation.

Detection and reconstruction methods have been proposed that generate dense reconstructions in the form of 3D cuboids \cite{ku2018joint} or meshes \cite{gkioxari2019mesh} directly from single images, without reliance on prior knowledge of object shapes. To make these systems more robust, researchers have also proposed fusing multiple observations from different viewpoints \cite{newcombe2011kinectfusion}, and coupling semantic information as priors into object reconstruction pipelines \cite{salas2013slam++,wang2021dsp}. Multi-view reconstruction methods \cite{whelan2015elasticfusion} generate 3D models from multiple frame, are also called structure from motion (SfM). Geometric priors have also been widely used for different objects and scenarios, e.g., shape prior~\cite{wang2021dsp,sucar2020nodeslam}, size prior~\cite{yang2019cubeslam,ok2019robust,liao2022so}. 
PointFlow~\cite{yang2019pointflow} uses normalization flow to learn a prior distribution of point clouds and reconstructs shapes from partial points.
It remains an open question to design an object representation with prior knowledge for detailed shape reconstruction from images and yet can generalize to many objects of different shapes.

\subsection{Neural Implicit Representation}

Recently, neural implicit representations have attracted wide attention in image and scene rendering \cite{mildenhall2020nerf,niemeyer2021giraffe}, voice encoding \cite{sitzmann2020implicit}, 3D objects~\cite{park2019deepsdf,mescheder2019occupancy} and scenes representations \cite{sucar2021imap}. 
For object representation, DeepSDF \cite{park2019deepsdf} proposes to use a neural network to approximate a continual signed distance function for modelling both known and unknown objects, which are captured via interpolation and completion of partial observation. Similar ideas are used in object reconstruction \cite{runz2020frodo, duggal2022mending}, object-level SLAM \cite{sucar2020nodeslam,wang2021dsp} and multiple object tracking \cite{li2021moltr}.
For example, Duggal et al. \cite{duggal2022mending} reconstruct cars from single-frame lidar points and optionally an image, but without uncertainty quantification and multi-view fusion.
Besides object-level details, researchers have further proved the effectiveness of neural implicit representations in representing large scenes, e.g., NeRF \cite{mildenhall2020nerf}, visual SLAM \cite{sucar2021imap,zhu2022nice,zhu2023nicer} and scene reconstruction \cite{sun2021neuralrecon}.

Implicit representation in 3D currently presents many open problems to address, such as effective neural architectures, multi-view fusion methods and uncertainty representations. As described above, this work focuses on identifying an effective fusion method and providing accurate uncertainty measures for downstream tasks.

\begin{figure*}[h]
\centering
\includegraphics[width=0.9\textwidth]{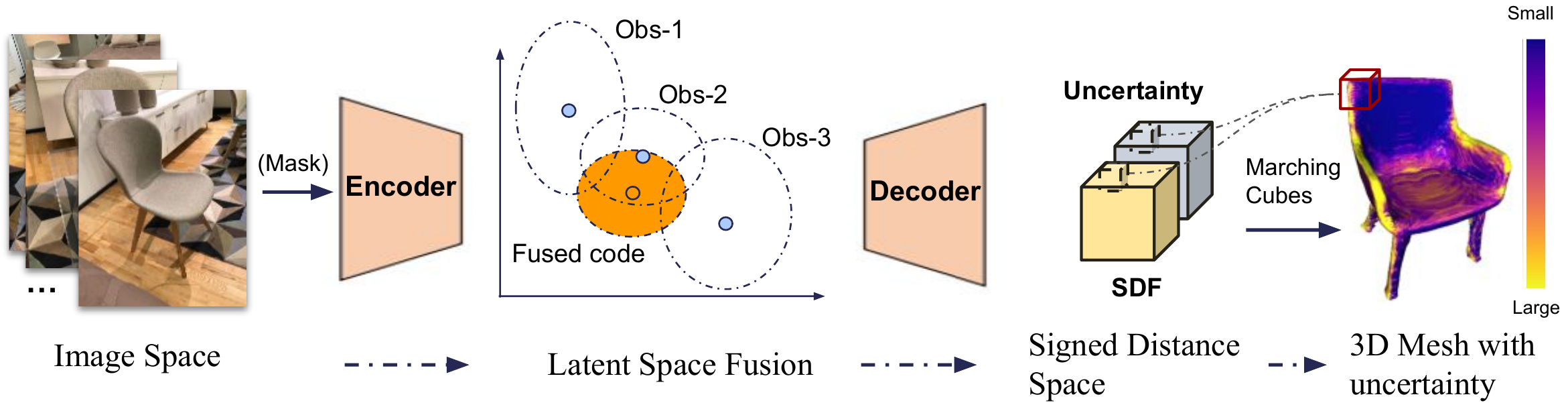}
\caption{Proposed System Framework. It consists of an uncertainty-aware image Encoder and a pre-trained decoder. We fuse multi-view observations in the latent space under a Bayesian framework. The decoder takes the fused latent space encoding and generates SDF values and associated uncertainties. Finally, the Marching Cubes algorithm is used to generate a mesh from the SDF values with uncertainty at each vertex. We visualize the relative uncertainty values with a color bar inside each models in this paper. }
\label{fig-framework}
\end{figure*}

\subsection{Uncertainty Modelling in Deep Learning}

Modelling uncertainty in deep learning inference has been well studied in the area of Bayesian Neural Networks~\cite{mackay1992practical}. Common uncertainty modelling techniques include sampling methods such as MC Dropout and Deep Ensembles, Error Propagation and Direct Modelling~\cite{feng2021review}. MC Dropout~\cite{gal2016uncertainty} and Deep Ensembles~\cite{lakshminarayanan2017simple} need to run the network multiple times to produce samples from which to estimate uncertainty. Directly Modelling \cite{kendall2017uncertainties} can output uncertainty from a single forward pass and is much more efficient, so we use it to estimate the uncertainty in our work. Error Propagation \cite{postels2019sampling} can also be run efficiently at inference time but requires complex modification of network layers which can affect network performance adversely, so we leave it as future work.

Direct modelling faces the problem of inaccurate and uncalibrated uncertainty in classification and regression \cite{kuleshov2018accurate}. Several methods are proposed to evaluate the output calibration, including calibration plot~\cite{guo2017calibration}, and proper scoring rules \cite{harakeh2021estimating} such as Energy Score and Negative Log Likelihood. A recalibration method \cite{guo2017calibration} has been proposed to rectify the calibration via temperature scaling. We will give a detailed analysis with proper scoring rules, and a calibration plot for our uncertainty output.

Very limited work exists for considering uncertainty in neural implicit representation. 
Researchers~\cite{deng2021deformed, duggal2022topologically, park2021hypernerf} have investigated learning a distribution of different topology shapes by changing a low dimensional hyperspace and can model the correspondences between shapes.
Deng et al.~\cite{deng2021deformed} models the correspondence uncertainty inside a shape category, instead of the reconstruction uncertainty from image observations, e.g., occlusion and ambiguity. It is also non-trivial to propagate the uncertainty for multi-view fusion.
Ours aims to derive from Bayesian formulation and output well-calibrated uncertainty.
Most related to ours is \cite{shen2022conditional} which models uncertainty in the color and density output of a scene-level neural representation. However, we concentrate on the problem of 3D object reconstruction and multi-view fusion. As far as we know, we are the first to estimate uncertainty for neural object representation from monocular images.

\section{Methods}

\subsection{Framework Overview}

The system framework is shown in Fig \ref{fig-framework}. The inputs are monocular image sequences of an object taken from different viewpoints. For each input image, the system outputs a reconstructed 3D object shape with uncertainty. The system can fuse multi-view observations in an uncertainty-aware way to incrementally update the shape. 

The system consists of an uncertainty-aware neural object representation, and an uncertainty-aware Image Encoder. The neural object representation learns an object shape prior in a latent code space. It has a decoder to generate Signed Distance Function (SDF) values conditioned on each latent code. Then, the Marching Cubes algorithm \cite{lorensen1987marching} is used to generate a mesh from the SDF values, with uncertainty represented as an isotropic variance attached to its vertices.

The uncertainty-aware Image Encoder takes in monocular images and outputs latent codes with uncertainty. 
In this work, we consider a diagonal covariance matrix for all the dimensions of the latent space. 
When there are multiple images, the multi-view fusion module fuses each output through a Bayesian update rule to estimate both the mean and covariance of the latent code. We now proceed with a more detailed formulation of our approach.

\subsection{Uncertainty-aware Neural Object Model}
\label{sec:uncer_propatgation}

Building on DeepSDF \cite{park2019deepsdf}, we propose to expand the current decoder-based neural object representation to model uncertainty. It is worth mentioning that the proposed uncertainty modelling and fusion method is generalizable to other similar neural representations with limited modification.

\textbf{3D object shape modelling with a neural network}. A neural network $f_\theta$ can be trained as a function to map any 3D coordinate, $\mathbf{X}=[x,y,z] \in \mathbb{R}^3$, to its SDF value of $s \in \mathbb{R}$:

\begin{equation} \label{eq:deepsdf_single}
s = f_\theta(\mathbf{X})
\end{equation}
where $\theta$ are the network parameters. Given a 3D grid of SDF values, the Marching Cubes algorithm can then generate a mesh. We can model a 3D shape with each parameter $\theta$. To model a specific class of objects, e.g. chairs or tables, we make the network conditional on a $D$-dimensional latent code, $\mathbf{z} \in \mathbb{R}^D$: 
\begin{equation} 
s = f_\theta(\mathbf{X}, \mathbf{z}), \mathbf{z} \in \mathbb{R}^D
\label{eq:deepsdf_multi}
\end{equation}
By varying $\mathbf{z}$, the SDF function will also change, as well as the 3D reconstruction it produces.  In this manner, a single decoder network can be trained to express the SDF representations of multiple semantically and geometrically similar objects, based on a latent code associated with each training object instance.

\textbf{Modelling uncertainty into 3D object shape.} In Eq. \ref{eq:deepsdf_multi}, the code $\mathbf{z}$ is deterministic. To model uncertainty, we model the $D$-dimensional latent code $\mathbf{z}$ as a probabilistic variable obeying a multivariate Gaussian distribution $\mathbf{z} \sim \mathcal{N}_D(\boldsymbol{\mu}, \boldsymbol{\Sigma})$. 
To simplify the problem, we assume each dimension of $\mathbf{z}$ is independent, which leads to a diagonal covariance matrix $\boldsymbol{\Sigma}$. We will train a neural network to output the mean and variance for each dimension of $\mathbf{z}$.

We also model the SDF value at $\mathbf{X}$ as a random variable, $s \sim \mathcal{N} (\mu_s, \sigma_s^2)$. According to Eq. \ref{eq:deepsdf_multi}, we can propagate the code uncertainty in $\mathbf{z}$ to the SDF value through the decoder network. Since the neural network $f_\theta$ is nonlinear, we can not directly solve for $\sigma_s^2$, and must employ some form or approximation to propagate the uncertainty from code input to SDF output. 

\textbf{Uncertainty propagation through neural network}. We use Monto Carlo Sampling \cite{hastings1970monte} to propagate the uncertainty through the nonlinear network. First, we sample $M$ codes $Z = \{\mathbf{z}_m\}_{m=1}^M$ from the code distribution $\mathbf{z} \sim \mathcal{N}_D(\boldsymbol{\mu}, \boldsymbol{\Sigma})$. For a 3D point $\mathbf{X}$, to get the variance $\sigma_s^2$ of its SDF, we pass each code $\mathbf{z}_m\in Z$ through Eq. \ref{eq:deepsdf_multi} to get $s_m$. We then calculate the sample variance \cite{chan1983algorithms} from the $M$ SDF values:

\begin{equation} \label{eq:sdf_net}
\sigma_s^2 = \frac{1}{M-1}{ \sum_{m=1}^M (s_m - s_\mu)^2}  
\end{equation}
where $s_\mu=\frac {1}{M} { \sum s_m}$ is the sample mean. We then calculate the SDF uncertainty for each of the vertices of the mesh generated using Marching Cubes. 
Now we can use the mean $\boldsymbol{\mu}$ and covariance matrix $\boldsymbol{\Sigma}$ of the latent code distribution $\mathcal{N}_D(\boldsymbol{\mu}, \boldsymbol{\Sigma})$ to represent a 3D object shape and its uncertainty.
The remaining question is how to estimate the mean $\boldsymbol{\mu}$ \textit{and} the covariance $\boldsymbol{\Sigma}$ from input images.

\subsection{Uncertainty-aware Image Encoder}
We propose training a simple encoder network $f_\beta$ to map an RGB image $\mathbf{m} \in \mathbb{R}^{H\times M\times3}$ with height $H$ and width $W$ to a D-dimensional latent code $\mathbf{z}$ with mean $\boldsymbol{\mu} \in \mathbb{R}^D$ and covariance $\boldsymbol{\Sigma} \in \mathbb{R}^{D \times D}$. Since we assume each code dimension is independent, the covariance matrix is diagonal and can be represented as $\boldsymbol{\Sigma}=diag(\boldsymbol{\sigma}^2)$, where $\boldsymbol{\sigma} \in \mathbb{R}^D$.

\begin{equation} 
\boldsymbol{\mu}, \boldsymbol{\sigma} = f_\beta(\mathbf{m})
\label{eq:encoder}
\end{equation}

We use the Direct Modelling~\cite{kendall2017uncertainties} approach to output uncertainty, which is well-established and does not add computational complexity. We leave the comparison of other uncertainty modelling methods as future work. The Encoder consists of a feature backbone, ResNet-50, and an output layer for the mean and variance. The architecture is straight forward and we concentrate on the choice of proper losses \cite{harakeh2021estimating} to generate calibrated and accurate uncertainty. We consider two common losses, Negative Log-Likelihood loss (NLL) and Energy Score. We conduct extensive experiments to explore the effectiveness compared with the baseline model trained without uncertainty. 
We will briefly introduce the two losses below. Their advantages and applications in object detection have been discussed in \cite{harakeh2021estimating}.

\textbf{NLL loss}. The NLL loss can be viewed as a standard L$_2$ loss weighted by uncertainty. Considering a batch of outputs $\{({\boldsymbol{\mu}}_i, {\boldsymbol{\sigma}}_i)\}_{i=1}^N$ directly from the encoder with $N$ data samples, and the ground-truth codes $\{\mathbf{z}_i\}_{i=1}^N$, NLL can be written as:

\begin{equation}
\mathrm{NLL}= \frac{1}{2N} \sum_{i=1}^N (\boldsymbol{\mu}_i-\mathbf{z}_i)^\intercal \boldsymbol{\Sigma}_i^{-1} (\boldsymbol{\mu}_i-\mathbf{z}_i) + \log {\det \boldsymbol{\Sigma}_i}
\end{equation}
where $\boldsymbol{\Sigma}_i=diag(\boldsymbol{\sigma}_i^2) \in \mathbb{R}^{D \times D}$ and $\boldsymbol{\sigma}_i \in \mathbb{R}^D$. The first term pushes down the error, where the variance, $\boldsymbol{\Sigma}_i$, acts to reduce the weight of samples in high uncertainty areas. The second, regularization term avoids uncertainty from growing too large.

\textbf{Energy Score}. Energy Score (ES) can be generalized to any distribution. It concentrates on optimizing the result of high uncertainty data samples to improve performance during training. For computation efficiency, we use a Monte-Carlo approximation version \cite{harakeh2021estimating}, which is represented as:

\begin{equation}
\begin{array}{l}\mathrm{ES}=\frac{1}{N} \sum_{i=1}^N\left(\frac{1}{M} \sum_{m=1}^M\left\|\mathbf{z}_{i, m}-{\mathbf{z}}_i\right\| \right. \\ 
\qquad \quad \left. -\frac{1}{2(M-1)} \sum_{m=1}^{M-1}\left\|{\mathbf{z}}_{i, m}-{\mathbf{z}}_{i, m+1}\right\|\right)
\end{array}
\end{equation}
where ${\mathbf{z}}_{i,m}$ is the $m^{th}$ i.i.d sample from $\mathcal{N}({\boldsymbol{\mu}}_i, {\boldsymbol{\Sigma}}_i)$. We take $M=1000$ with very little computational overhead.

\subsection{Multi-view Bayesian Fusion in Latent Space}

\textbf{Bayesian Fusion}. 
Consider $N$ input images and the corresponding outputs $\{({\boldsymbol{\mu}}_i, {\boldsymbol{\sigma}}_i)\}_{i=1}^N$ from the encoder. Since we assume each of the $D$ code
dimensions is independent, the covariance matrix is diagonal $\boldsymbol{\Sigma}_i=diag(\boldsymbol{\sigma}_i^2)$.
We follow Gaussian Inference~\cite{barfoot2017state} to get a fused latent code $\mathbf{z}$. It follows the Gaussian distribution $\mathbf{z} \sim \mathcal{N}_D(\boldsymbol{\mu}, \boldsymbol{\Sigma})$, where:

\begin{equation}
\boldsymbol{\mu}=\boldsymbol{\Sigma} \sum_{i=1}^N \boldsymbol{\Sigma}_i^{-1} \boldsymbol{\mu}_i, \quad \boldsymbol{\Sigma}=\left(\sum_{i=1}^N \boldsymbol{\Sigma}_i^{-1}\right)^{-1}
\label{eq:gaussian_fusion}
\end{equation}
Then, we can use Monto Carlo Sampling to propagate $\mathbf{z}$ through the decoder to get the mean and variance of the SDF value for each 3D point as described in Sec. 3.2.

\textbf{Outlier rejection}. When facing extreme situations such as highly occluded objects, experimentation revealed that performance improves by treating them as outliers and filter them out of the fusion process, instead of incorporating them with high uncertainty. We define a modified inference strategy, ``Bayesian-$N$", which only selects the $N$ observations with the lowest uncertainty for Bayesian fusion. 
When $N=1$, we simply select the lowest uncertainty viewpoint. When $N=N_{max}$, we use all available measurements without rejection, referred to as ``Bayesian'' by omitting $N$.

\section{Experiments}

\begin{table}[]
\begin{tabular}{@{}llccc@{}}
\toprule
\multicolumn{1}{c}{Methods} & \textbf{Shape} & \textbf{IoU} $\uparrow$   & \textbf{EMD} $\downarrow$   & \textbf{CD} $\downarrow$   \\ \midrule
3D-R2N2~\cite{choy20163d}                     & Voxel                           & 0.136       & 0.211       & 0.239        \\
PSGN~\cite{fan2017point}              & Points                          & N/A                             & 0.216                           & 0.2                             \\
3D-VAE-GAN~\cite{wu2016learning}                  & Voxel                           & 0.171       &0.176       & 0.182       \\
DRC~\cite{tulsiani2017multi}               & Voxel                           & 0.265                           & 0.144                           & 0.16                            \\
MarrNet~\cite{wu2017marrnet}                     & Voxel                           & 0.231                           & 0.136                           & 0.144                           \\
AtlasNet~\cite{groueix2018papier}          & Mesh                            & N/A                             & 0.128                           & 0.125                           \\
Sun et al.~\cite{sun2018pix3d}                  & Voxel                           & 0.282                           & 0.118                           & 0.119                           \\
% Pix2Vox++ (IJCV’20)         & Voxel                           & 0.292                           & N/A                             & N/A                             \\
FroDO~\cite{runz2020frodo}             & Mesh                            & 0.302                           & 0.112                           & 0.103 \\ \midrule
FroDO* w/ GT Mask           & Mesh                            & 0.319                           & 0.107                           & 0.109                           \\
FroDO* w/ Seg Mask          & Mesh                            & 0.285                           & 0.120                            & 0.121                           \\
FroDO* w/o Mask             & Mesh                            & 0.257                           & 0.127                           & 0.123                           \\ \midrule
Ours w/ GT Mask             & Mesh                            & \textbf{0.335} & \textbf{0.102} & \textbf{0.102}                           \\
Ours w/ Seg Mask            & Mesh                            & 0.293                           & 0.116                           & 0.116                           \\
Ours w/o Mask               & Mesh                            & 0.268                           & 0.122                           & 0.118                           \\ \bottomrule
\end{tabular}

\caption{Single-view reconstruction of the chairs category on the Pix3D dataset. Metrics include Intersection of Union (IoU), Earth Moved Distance (EMD) and Chamfer Distance (CD). GT: Groundtruth. Seg: Semantic Segmentation algorithm. * Our own implementation.}

\label{table:pix3d-monocular}

\end{table}

\begin{figure}[t]
\centering
\includegraphics[width=0.48\textwidth]{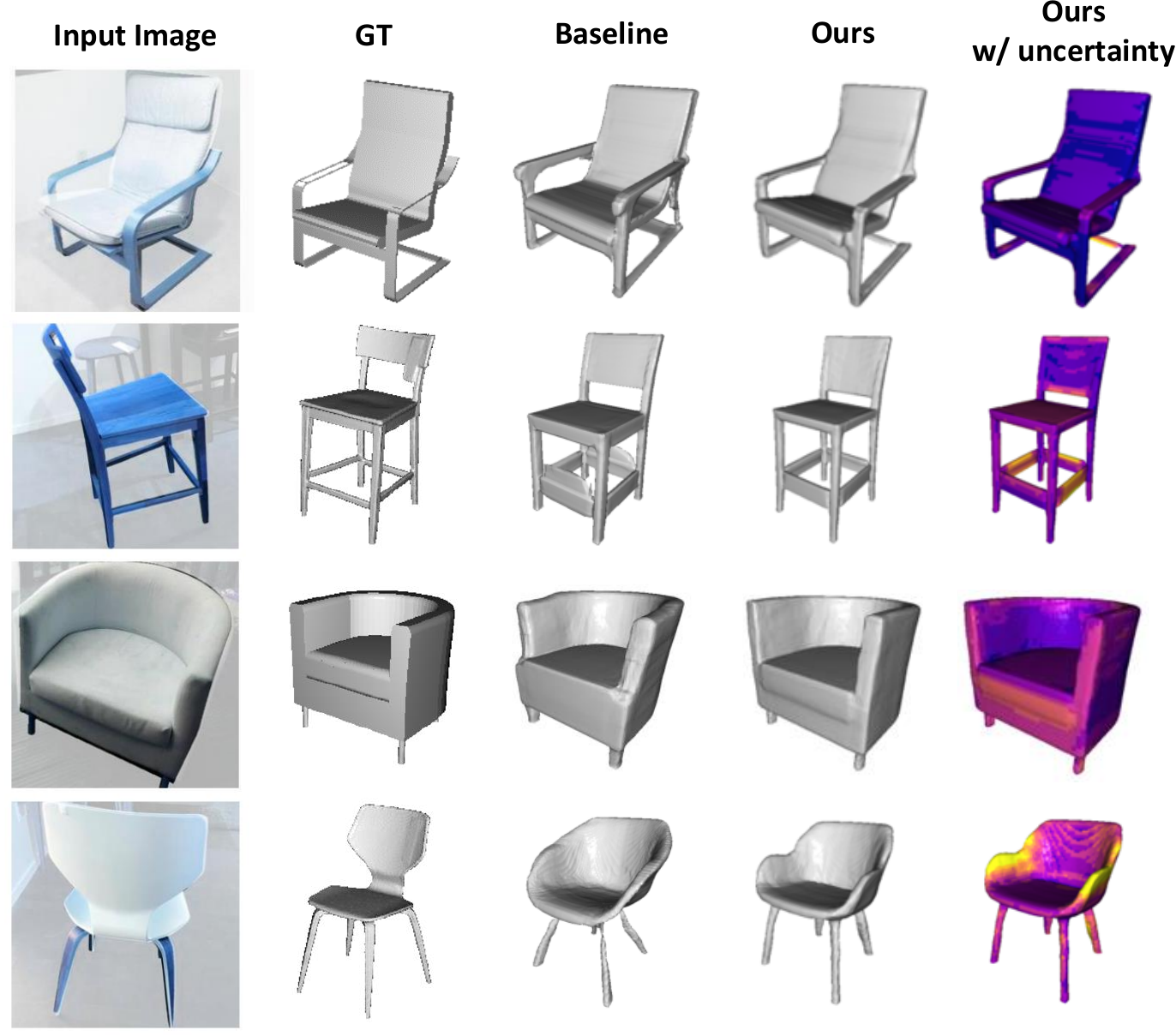}
\caption{Qualitative results of single-view reconstruction on chairs of Pix3D dataset. The models are trained on ShapeNet dataset. Ours has fewer artifacts than the baseline FroDO*, and can further output uncertainty for each object part. Our uncertainty highlights the areas with errors, indicating information for downstream tasks.}
\label{fig:pix3d_mono}
\end{figure}

\begin{figure}[t]
\centering
\includegraphics[width=0.48\textwidth]{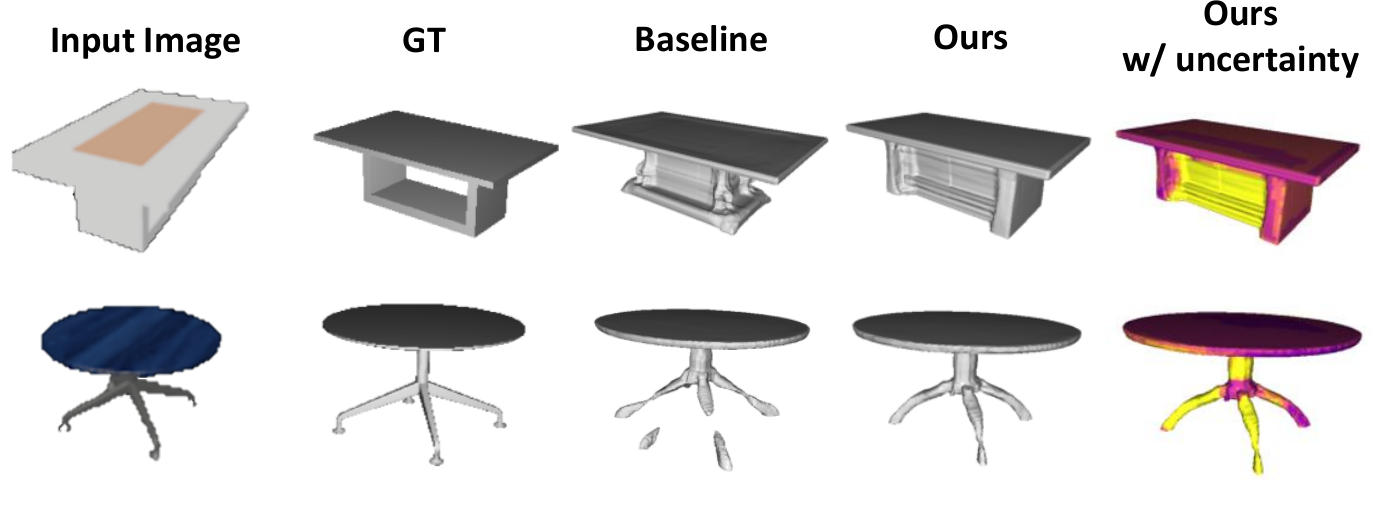}
\caption{Qualitative results of single-view reconstruction of tables on ShapeNet dataset. The models are both trained and tested on ShapeNet dataset. Compared with the baseline Frodo*, Ours further outputs uncertainty and has fewer artifacts.}
\label{fig:shapenet_mono}
\end{figure}

\begin{figure*}[t]
\centering
\includegraphics[width=1.0\textwidth]{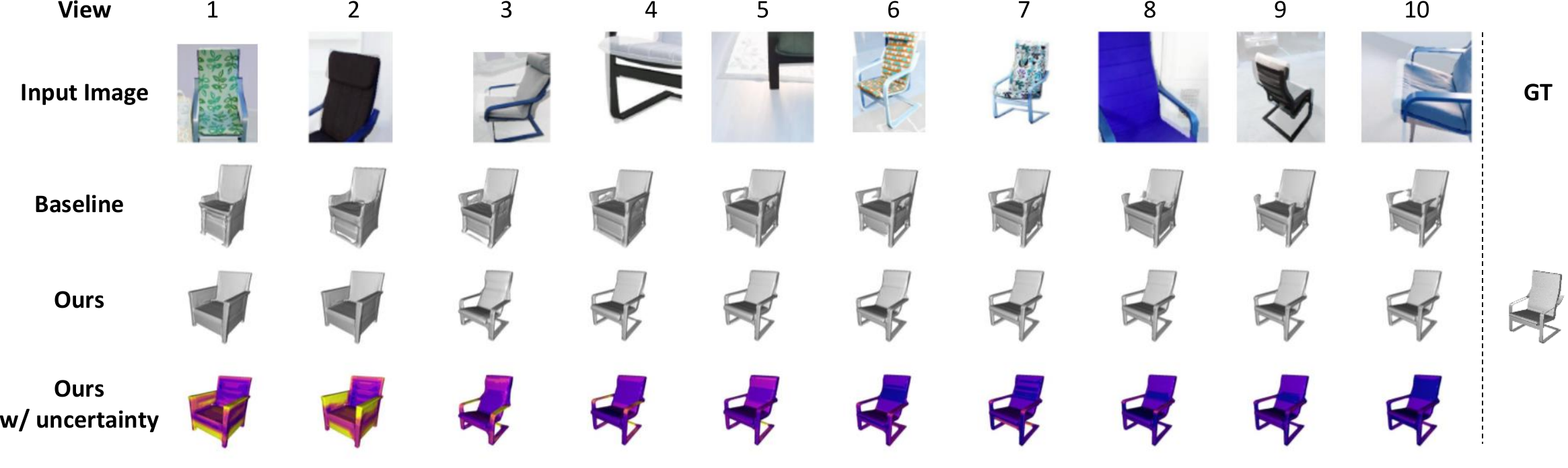}
\caption{Qualitative results of multi-view fusion on the Pix3D dataset. After fusing 1 to 10 corrupted observation images, Ours reduces the shape uncertainty based on Bayesian Fusion and outputs a more accurate final reconstruction compared with the baseline FroDO*. Note that our method can fuse observations without knowing the camera poses and can work even when the objects have different textures.}
\label{fig:vis-fusion}
\end{figure*}

\subsection{Implementation and Training Details}

Our system consists of an encoder and a decoder. For the decoder, we follow the implementation and training of DeepSDF \cite{park2019deepsdf} on ShapeNet \cite{chang2015shapenet}. 
For the encoder, we use ResNet-50 pretrained on ImageNet as the feature backbone, modify the output layer to the dimensions of the code $N$, and further add $K$ dimensions for the uncertainty. We take $N=K=64$ in the experiments. 

We need monocular images and ground-truth latent codes to train the encoder. We use the images from ShapNetRendering dataset~\cite{choy20163d} which contains rendered images of 24 different views from the CAD models in ShapeNet \cite{chang2015shapenet}. After training the decoder, we get optimized latent codes for each CAD models, and we use them as the ground-truth latent codes for the training and evaluation of the Encoder.

For training the Encoder, we use the same dataset split as FroDO~\cite{runz2020frodo}.
We augmented the training data with random resize and horizontal flip, and random background clip from SUN dataset~\cite{xiao2010sun}. 
We set a learning rate of 0.1, a batchsize of 64, and a random seed of 1000. We use a polynomial learning rate scheduler, and trained for 50 epochs. 

To verify that our model can generalize to different categories, we test on both chairs and tables categories on ShapeNet dataset. To verify the performance in real scenarios, after training on the sythetic Shapenet dataset, we directly evaluate on the Pix3D dataset without finetunning.

\subsection{Metrics and Baselines}

\textbf{Metrics}. For the reconstruction, we calculate the metrics of Intersection over Union (IoU), Chamfer Distance (CD), and Earth Moved Distance (EMD) on the voxelized mesh with a resolution of $32^3$ following~\cite{sun2018pix3d}.
For the uncertainty, we use Negative Log Likelihood (NLL) and Energy Score (ES), which evaluate the error of the regression, and the calibration and sharpness of the estimated uncertainty.

\textbf{Baselines}. We train our model with Energy Score and denote it as Ours. We also compare the choice of the two training losses, Energy Score and NLL in the ablation study.

FroDO~\cite{runz2020frodo} is a baseline closest to ours with an encoder trained with L2 loss and a DeepSDF decoder for reconstruction but without uncertainty. It averagely fuses multiple latent codes to get the final reconstruction. 
It also supports pose estimation and optimization with both shape and pose together. 
Since pose estimation is out of the scope of the paper, we compare with the Encoder parts to investigate the effectiveness of uncertainty. Note that the results on Pix3D dataset of the origin paper do not use the pose module so it is a fair comparason.
Since FroDO is not open-sourced, to fully investigate the performance, we implemented it by ourselves and denote it as FroDO*. 
We compare our implemented version with origin published version on Pix3D dataset.
We also fully compare our models with other published state-of-the-art models for the reconstruction accuracy on Pix3D dataset.

\subsection{Single-view reconstruction}

\begin{table}[]
\centering
\begin{tabular}{@{}l|c|l|lr@{}}
\toprule
                        & \multicolumn{1}{l|}{\textbf{Views}} & \textbf{Methods} & \textbf{IoU}    & \multicolumn{1}{l}{\textbf{\%}} \\ \midrule
\multirow{2}{*}{FroDO*} & 1                                   & Single-view      & 0.3225          & 0                               \\ \cmidrule(l){2-5} 
                        & 10                                  & Average          & 0.3456          & 0                               \\ \midrule
\multirow{12}{*}{Ours}  & 1                                   & Single-view      & 0.3373          & 4.6                             \\ \cmidrule(l){2-5} 
                        & \multirow{11}{*}{10}                & Average          & 0.3750          & 8.5                             \\ \cmidrule(l){3-5} 
                        &                                     & Bayesian-1       & 0.3719          & 7.6                             \\
                        &                                     & Bayesian-2       & 0.3828          & 10.8                            \\
                        &                                     & Bayesian-3       & 0.3874          & 12.1                            \\
                        &                                     & Bayesian-4       & \textbf{0.3902} & \textbf{12.9}                   \\
                        &                                     & Bayesian-5       & 0.3867          & 11.9                            \\
                        &                                     & Bayesian-6       & 0.3867          & 11.9                            \\
                        &                                     & Bayesian-7       & 0.3836          & 11.0                            \\
                        &                                     & Bayesian-8       & 0.3828          & 10.8                            \\
                        &                                     & Bayesian-9       & 0.3822          & 10.6                            \\
                        &                                     & Bayesian(-10)      & 0.3816          & 10.4                            \\ \bottomrule
\end{tabular}
\caption{Multi-view fusion IOU performance on the chair category of the Pix3D-MV dataset. Ours with uncertainty can higher IOU than the deterministic baseline. $\%$ denotes percent improvement over baseline single/multiview. }
\label{table:pix3d_mv_uncertain_bestk}
\end{table}

\begin{table}[]
\centering

\resizebox{1.0\columnwidth}{!}{%

\begin{tabular}{@{}lrrrrr@{}}
\toprule
                 & \multicolumn{5}{c}{\textbf{Min Scale}}                                             \\ \midrule
\textbf{Methods} & \textbf{1.0}     & \textbf{0.8}   & \textbf{0.4}   & \textbf{0.2}   & \textbf{0.1}   \\ \midrule
FroDO*            & 0.346          & 0.343          & 0.338          & 0.323          & 0.318          \\ \midrule
NLL - Bayesian   & 0.327          & 0.326          & 0.318          & 0.302          & 0.301          \\
NLL - Bayesian-4     & 0.346          & 0.349          & 0.335          & 0.330          & 0.341          \\ \midrule
Ours - Bayesian  & 0.382          & 0.362          & 0.345          & 0.332          & 0.322          \\
Ours - Bayesian-4    & \textbf{0.390} & \textbf{0.381} & \textbf{0.374} & \textbf{0.369} & \textbf{0.365} \\ \bottomrule
\end{tabular}

} % resize box

\caption{Multi-view reconstruction (IoU) on the Pix3D-MV chair set when the input images are cropped to a randomly selected area between [Min Scale, 1.0]. As Min Scale decreases, the fusion task becomes more difficult. Ours with uncertainty gets better robustness in difficult tasks.}
\label{table:pix3d_manual_challenge}
\end{table}

We show the results on the chairs category of Pix3D dataset in Fig.~\ref{fig:pix3d_mono} and Table \ref{table:pix3d-monocular}. 
When using the origin RGB image as input (see Ours w/o Mask), Ours outperforms the baselines PSGN, DRC, AtlasNet and our implementation of Frodo, but still has a gap with the published state-of-the-art performance from Frodo.
We trained the Encoder on the synthetic dataset ShapeNet and inferred on the real Pix3D dataset. The domain gap of the real texture and background limits the performance of the encoder, which is only a vanilla ResNet originally designed for classification. 

We further use Mask2Former, an off-the-shelf semantic segmentation method~\cite{cheng2021mask2former}, to filter the background (see Ours w/ Seg Mask), and notice an obvious improvement on the reconstruction accuracy. When using groundtruth masks to filter all the background (see Ours w/ GT Mask), ours achieves an IoU of 0.335 and outperforms all the baselines, which demonstrates the accuracy upper bound of our design of an image encoder and a shape decoder. It demonstrates that the decoder learns a powerful category-level prior distribution of the objects shape even from the synthetic dataset, and can transfer to real objects.

Further training and fine-tuning with background augmentation, e.g., randomly sampled texture images, has been shown to make the network adapt to varied backgrounds~\cite{shorten2019survey}. However, the evaluation of the background invariance is out of the scope of this paper. 

We concentrate on the performance improvement brought by the uncertainty both in single-view and multi-view scenarios. We implemented the Encoder in FroDO~\cite{runz2020frodo} which outputs deterministic latent code as baseline, and keep the same decoder. We show the performance when using different masks (w/o Mask, w/ Seg Mask and w/GT mask). Ours gets better performance for all the mask types, thanks to the introduction of uncertainty during training which makes the network more robust to the domain gap so that it does better on difficult reconstruction tasks.

Our implementation (FroDO w/o Mask) has a moderate gap compared with the published result of FroDO. 
However, when using GT mask (FroDO w/ GT Mask), it achieves higher performance than the published result, which indicates the validity of the accuracy upper bound. 
Despite extensive experiments in data augmentation and training methods, we were unable to recreate the published performance of Frodo, and the authors of that work have not released their code.
In the following experiments, we highlight the effectiveness of uncertainty during multi-view fusion by comparing our work to our own implementation of FroDO*.

\subsection{Multi-view Reconstruction}

The Pix3D dataset contains real images and groundtruth CAD models but has no splits for instances and their multi-view observations. To evaluate the multi-view performance, we group the images of the chair category into separate instances according to their GT models, and keep 10 views as one instance, which results in a multi-view dataset with totally 1490 images from 149 instances. We denote this multi-view dataset as Pix3D-MV which is a subset of the original Pix3D dataset.
% 1. multi-view reconstruction compared w/ no uncertainty
We show the results of multi-view fusion on Pix3D-MV chair set in Fig.~\ref{fig:vis-fusion} and Table~\ref{table:pix3d_mv_uncertain_bestk}. 
We consider the following methods as multi-view fusion baselines: \textit{Average} equally fuses each estimated latent code; \textit{Bayesian} fuses with uncertainty according to Bayesian Fusion in Equation~\ref{eq:gaussian_fusion}; \textit{Bayesian-$K$} keeps the top-$K$ observations with the lowest uncertainty evaluated by taking the trace of the covariance matrix, and then fuses with \textit{Bayesian}.

Compared with the deterministic baseline, Ours with uncertainty achieves an IoU of 0.3816 vs. 0.3456  with a margin of 10.4\%. When using Bayesian-4 to filter the outliers and keep the first 4 observations, ours can further boost up to an IoU of 0.3902 with a margin of 12.9\% compared with the baseline. The experiment demonstrates that uncertainty can effectively help the system to trust the observations that contain more valid information, and make the system more robust to outliers in the multi-view observations.

We further push the limit of the robustness brought by the uncertainty during multi-view fusion in Table~\ref{table:pix3d_manual_challenge}. In real applications like robots, the input images are heavily corrupted because of occlusions, errors in segmentation or sensor noise. We simulate challenging situations by randomly cropping images into a specific size range [$c$, 1.0], so that only part of the origin image is kept. By changing the value of the min scale, $c$, we vary the difficulty of the experiments. As is visible in Table~\ref{table:pix3d_manual_challenge}, when the min scale decreases, the task becomes more difficult. In the most difficult task, where images can be cropped to only 10\% of the origin images, the deterministic FroDO model suffers from the occlusions obviously and decreases to an IoU of only 0.318 while Ours w/ uncertainty remains robust to the cropping and maintains an IoU of 0.366, with an improvement of 15.1\%. The experiments prove the effectiveness of using uncertainty in multi-view fusion to select valid information from a group of corrupted input images.

\subsection{Ablation Study}

\textbf{Loss function.} We compare two options for uncertainty training loss: NLL (NLL) and ES (Ours) in Table~\ref{table:pix3d_manual_challenge}. Even though training with NLL can improve the performance in difficult tasks, it presents lower accuracy in general than when training with ES.  
% Explanation of ES is better than NLL

\textbf{Selection of $K$ in Bayesian.} 
With uncertainty, we can detect outliers and take active actions to deal with them. 
As in Table~\ref{table:pix3d_mv_uncertain_bestk}, when decreasing $K$, the performance increases since the outlier codes are filtered out. The highest IOU performance of 0.39 is achieved with $k=4$, which has an improvement of 12.9\% compared with the baseline. When further decreasing the number, the system has too few observations to fuse and the performance begins to drop. An interesting finding is that, with 1 best views we get better performance than the \textit{Average Equal} of FroDO. In real applications, we have the option of adjusting the parameter $K$ to better suit the data. 

\subsection{Evaluation on ShapeNet}

\begin{table}[]
\centering

\resizebox{1.03\columnwidth}{!}{%

\begin{tabular}{llcclcrl}
\hline
\multicolumn{1}{c}{}                                   &                         & \multicolumn{3}{c}{Chairs}                                                                       & \multicolumn{3}{c}{Tables}                                                                       \\ \cline{3-8} 
\multicolumn{1}{c}{\multirow{-2}{*}{\textbf{Methods}}} & \multirow{-2}{*}{Views} & Easy           & Mid                                    & Hard                                   & Easy           & \multicolumn{1}{c}{Mid}                & Hard                                   \\ \hline
FroDO*                                                 &                         & \textbf{0.388} & 0.341                                  & 0.326                                  & 0.403          & 0.321                                  & 0.300                                  \\
Ours                                                   & \multirow{-2}{*}{1}     & 0.383          & \textbf{0.343}                         & 0.326                                  & \textbf{0.410} & 0.321                                  & 0.300                                  \\ \hline
FroDO*                                                 &                         & \textbf{0.410} & \cellcolor[HTML]{EFEFEF}0.375          & \cellcolor[HTML]{EFEFEF}0.359          & 0.446          & \cellcolor[HTML]{EFEFEF}0.376          & \cellcolor[HTML]{EFEFEF}0.350          \\
Ours                                                   & \multirow{-2}{*}{10}    & 0.400          & \cellcolor[HTML]{EFEFEF}\textbf{0.391} & \cellcolor[HTML]{EFEFEF}\textbf{0.385} & 0.446          & \cellcolor[HTML]{EFEFEF}\textbf{0.399} & \cellcolor[HTML]{EFEFEF}\textbf{0.378} \\ \hline
\end{tabular}

} % resize box

\caption{Single- and multi-view reconstruction results of the Chairs and Tables on the ShapeNet dataset. Ours shows obvious improvements when the task becomes hard for multi-view fusion, showing the robustness to corruption brought by uncertainty.}
\label{table:shapenet}
\end{table}

Our network architecture, including the uncertainty framework, is not specifically designed for any categories. If the training data is available, we can support the new categories.  We show more experiments results on the Chairs and Tables categories on ShapeNet dataset in Table~\ref{table:shapenet}. We also show the results of Tables in Fig.~\ref{fig:shapenet_mono}. We train each categories separately.  During inference, we consider two tasks, \textit{Easy} for taking origin rendered images in Shapenet, and \textit{Mid}/\textit{Hard} for randomly croping the images into a range of [$c$,1.0] ($c=0.1$ for \textit{Mid} and $c=0.01$ for \textit{Hard}). For the multi-view fusion method, we use \textit{Bayesian Fusion} for Ours and \textit{Average} for FroDO*.
In the Easy task, we get comparable but slightly lower IOU performance on Chairs, but higher IoU performance on Tables. Since our model, with the same architecture, requires a part of the model capacity to regress uncertainty. It is notable that uncertainty is not very helpful in easy tasks where each input image contains enough information for reconstruction. 
In the Mid and Hard task, the effectiveness of uncertain becomes more pronounced, as Ours gets higher multi-view accuracy than the baseline. Especially on Hard task, we got an IoU of 0.385 v.s. 0.359 on Chairs, and 0.378 v.s. 0.350 on Tables.
This result demonstrates that uncertainty can robustly find and fuse the valid data from a set of input data of varying quality.

\section{Conclusion}

We propose an uncertainty-aware 3D object reconstruction framework that can take in both monocular and multi-view images. 
Based on the neural shape models, we introduce a method to model and estimate uncertainty in latent space and a method to propagate uncertainty into 3D object space, so that we can output 3D object shape with uncertainty awareness.
Our proposed method can be trained on a purely synthetic dataset and directly evaluated on real datasets. It achieves 
higher reconstruction performance than deterministic models, and in particular demonstrates better robustness and accuracy in multi-view fusion when the input image sequences are corrupted. 

In future work, we plan to scale up to multi-classes objects reconstruction and uncertainty estimation. Also, it will be interesting to leverage the uncertainty-aware shape model for down stream tasks related to objects such as detection, segmentation, tracking, and object-level SLAM.

\section*{Acknowledgement}

The authors would like to thank Jordan Hu for the fruitful discussion on uncertainty.

{\small
\bibliographystyle{ieee_fullname}
\bibliography{egbib}
}

\newcommand{\beginsupplement}{
  \setcounter{table}{0}  
  \renewcommand{\thetable}{S\arabic{table}} 
  \setcounter{figure}{0} 
  \renewcommand{\thefigure}{S\arabic{figure}}
  \setcounter{section}{0}
  \renewcommand{\thesection}{S\arabic{section}}
}

\beginsupplement

\section*{Supplementary Materials}

\section{Uncertainty in Downstream Tasks}

We have carried out experiments to prove the effectiveness of uncertainty-aware shape models in reconstruction and multi-view fusion. Besides, our work opens up the possibilities for multiple downstream tasks which we defer to future work. 

\begin{figure}[t]
\centering
\includegraphics[width=0.48\textwidth]{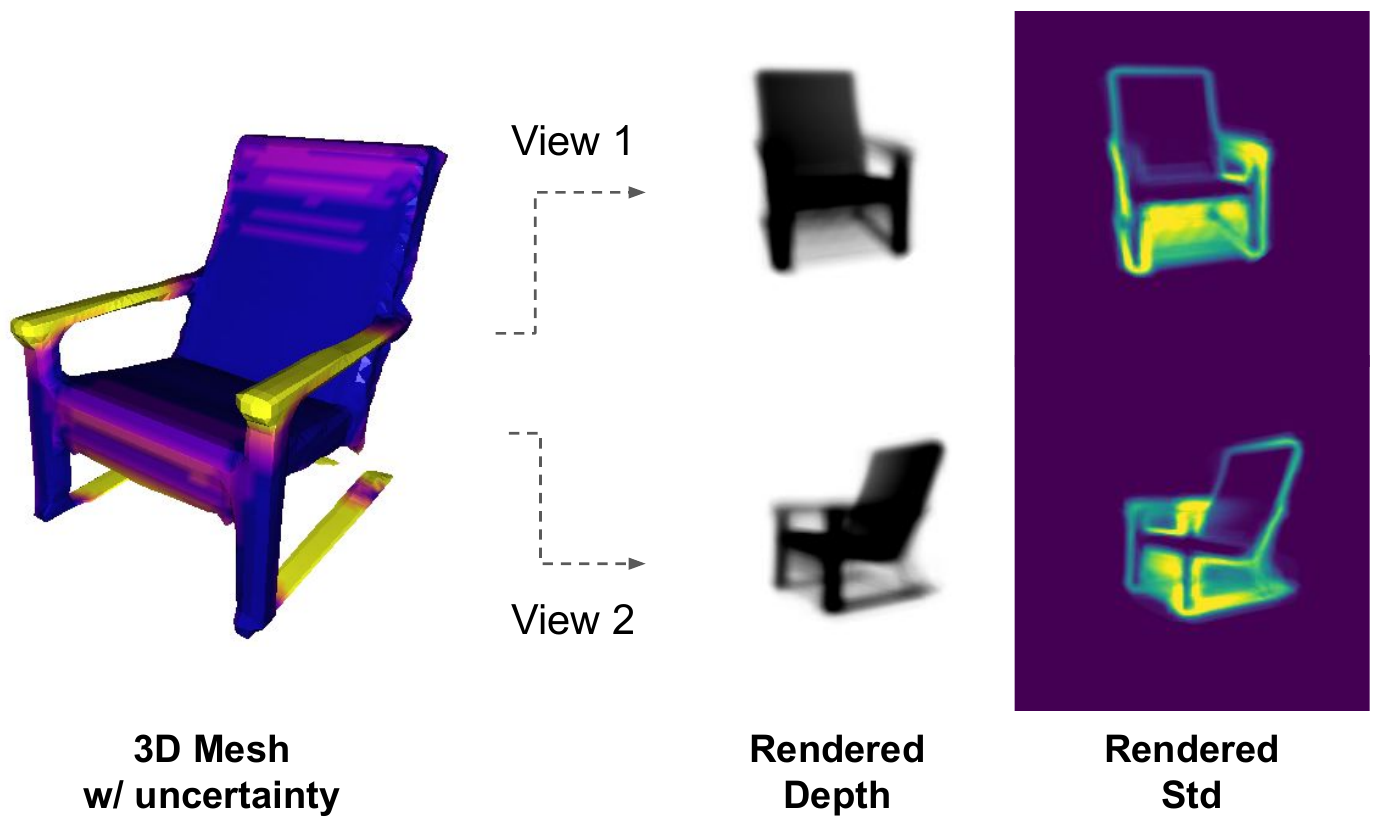}
\caption{Differentiable rendering with uncertainty. We can propagate the uncertainty in 3D shapes into any camera view with differentiable rendering. The rendered uncertainty (std) can be used by downstream tasks, such as constructing losses and propagating gradients for pose and shape optimization and object-level SLAM.}
\label{fig:render_2d}
\end{figure}

\textbf{Uncertainty for perception}. Since our uncertainty-aware shape model is continuous and differentiable, we can propagate the uncertainty stored in the shape into any observation model with a given math formulation. 
For example, differentiable rendering~\cite{kato2020differentiable} is the core technology to constrain neural implicit representation~\cite{mildenhall2020nerf, park2019deepsdf}, and solve perception tasks such as camera pose estimation~\cite{yen2020inerf}, object shape and pose estimation~\cite{bruns2022sdfest}, and object-level SLAM~\cite{wang2021dsp,sucar2020nodeslam}. They are the core abilities for robotics, VR \& AR, and autonomous driving. 
In Figure ~\ref{fig:render_2d}, we show a simple example of propagating the uncertainty in the 3D shape model into any given camera view with differentiable rendering. We sample multiple latent codes and calculate the sample mean and variance of the rendered depth of each latent code. Our rendered uncertainty map can be used in the differentiable constraints, which naturally assigns weights for areas with high information, to fuse multi-view observations in challenging real scenarios with occlusion and ambiguity.

\textbf{{Uncertainty for planning and control}}.
For indoor robots or autonomous driving cars operating in real scenarios, their world model (e.g., a map of objects) is uncertain with partial, noisy, and limited observations. 
Our model explicitly quantifies the uncertainty and stores it in the shape model, which can be used by the downstream modules to make decisions.
In motion planning~\cite{pairet2021online}, uncertainty helps safely navigate and stay away from uncertain objects to avoid obstacle. 
In robotic grasping~\cite{jin2023neu}, 
uncertainty can efficiently guide the manipulator to find a next-best-view to better reconstruct the uncertain part of the shape to guarantee the success rate of the grasps. Our model has the potential to offer an uncertainty source for those tasks.

\section{Uncertainty Analysis}

\subsection{Calibration Plot}
\label{sec:uncertain_analysis_calibration_plot}

\begin{figure}
     \centering
     
     \includegraphics[width=0.45\textwidth]{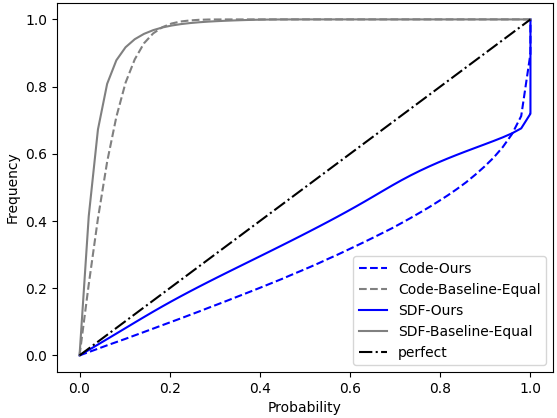}
     \caption{The calibration plot on the Pix3D dataset. The plots for the uncertainty of the latent space and SDF space are shown separately. We further show a baseline outputting equal uncertain of 1. Our approach is siginficantly better calibrated (closer to $Y=X$ line) than the baseline.}

     \label{fig:calibplot}
\end{figure}

We have evaluated the effectiveness of using uncertainty on reconstruction and multi-view fusion. 
We further evaluate the estimated uncertainty quality of correlating with the accuracy.
An estimated uncertainty is \textit{well-calibrated} when the estimation errors are lying into the predicted uncertain threshold. 
We introduce how we draw the calibration plot in Sec~\ref{sec:draw_calibration_plot}.

We present the calibration plot on the Pix3D dataset in Figure~\ref{fig:calibplot}, where the $X$ axis shows the predicted probability, and the $Y$ axis counts the real frequency where the estimation errors lying within the predicted uncertain threshold. 
A well-calibrated plot should be close to the line $Y=X$, where the real frequency aligns perfectly with the predicted probability.
Our method outputs uncertainty in latent space and the signed distance space, as in Figure~2.
The uncertainty in the latent space is directly output from the Encoder, and then propagates using Monte-carlo sampling through the decoder to the signed distance space, as described in Sec~3.2.
We draw calibration plots for both of them. We use a baseline with an equal variance of 1 for each code dimension. 
We can see that it is non-trivial to output a well-calibrated uncertainty for both the latent code and the 3D SDF values.
The calibration plots of the baseline method are far away from the line of $Y=X$, while ours is significantly closer, which shows a much better calibrated result than the baseline. 
At the same time, we still see an opportunity for further improvement, e.g., with temperature scaling~\cite{guo2017calibration}. We hope our approach can serve as a benchmark for future research that can output better uncertainty for 3D neural shape models.

\subsection{Drawing a Calibration Plot}
\label{sec:draw_calibration_plot}
We present a calibration plot in Sec~\ref{sec:uncertain_analysis_calibration_plot}. 
We describe in detail here the method to draw a calibration plot for the latent space and the signed distance function space.
Given $N$ 1-dimensional random variables $\{X_i\}^N_{i=1}$, for each $X_i$, our algorithm outputs the estimated univariate Gaussian distributions with the mean and variance of $\{(\mu_i, \sigma_i^2)\}_{i=1}^{N}$. 
To draw a calibration plot, we first uniformly sample $T$ probability $p_t$ in the range of $(0,1)$. For each $p_t$, we count the real frequency $F_t$ that the estimation errors $\{|X_i-\mu_i|\}_{i=1}^N$ lying inside the predicted threshold $r_i^t$:
\begin{equation}
F_t = \frac{1}{N}\Sigma_{i=1}^N I_{|X_i-\mu_i| \leq r_i^t}
\label{eq:frequency_calibration_plot}    
\end{equation}
where $I_C$ is an indicator function outputting 1 when $C$ holds, and 0 otherwise.
The predicted threshold $r_i^t$ can be calculated using the quantile function, which is the inverse cumulative distribution function.
For a normal distribution, the cumulative distribution function calculates the probability $p$ of a random variable $X$ lies inside a given range $x$:
\begin{equation}
F_X(x):=Pr(|X| \leq x)=p
\end{equation}
The quantile function inversely calculates the range of $x$ that the variable $X$ lies inside with a given probability $p$:
\begin{equation}
Q(p)=F_X^{-1}(p)=x
\end{equation}
For a normal distribution, the value of $Q(p)$ can be directly queried given any $p$. For example, $Q(0.9984)\approx3$ for a normal Gaussian, which is the 3-$\sigma$ principle.
For a Gaussian distribution with mean and variance of $(\mu_i, \sigma_i^2)$, we can calculate the predicted threshold $r_i^t=Q(p_t)\sigma_i$ by multiplying the standard deviation. Then we can count the real frequency $F_t$ according to Eq.~\ref{eq:frequency_calibration_plot}.
Finally, we can draw a plot with the sampled probability $\{p_t\}_{t=1}^T$ as the $X$ axis, and the counted frequency $\{F_t\}_{t=1}^T$ as the $Y$ axis. When drawing the calibration plot, we consider each dimensions of each latent codes independently, and also consider each SDF values independently.

\subsection{Metric Evaluation}

\begin{table*}[]
\centering
\begin{tabular}{@{}c|cc|cc|cc|c@{}}
\toprule
\multicolumn{1}{l|}{} & \multicolumn{2}{c|}{All Space}    & \multicolumn{2}{c|}{Surface}      & \multicolumn{2}{c|}{Non-Surface}  & Reconstruction \\ \midrule
\textbf{Methods}      & ES $\downarrow$              & NLL $\downarrow$             & ES $\downarrow$              & NLL $\downarrow$             & ES $\downarrow$              & NLL $\downarrow$             & IoU $\uparrow$            \\ \midrule
Equal Var             & 0.1932          & -0.203          & 0.1914          & -0.210          & 0.1939          & -0.201          & /              \\
Ours-NLL                   & 0.0353          & -2.720          & 0.0235          & -3.065          & 0.0399          & -2.587          & 0.299          \\
Ours-ES                  & \textbf{0.0338} & \textbf{-2.753} & \textbf{0.0221} & \textbf{-3.095} & \textbf{0.0384} & \textbf{-2.621} & \textbf{0.335} \\ \bottomrule
\end{tabular}
\caption{Uncertainty analysis in the 3D SDF space. Ours trained with ES shows better performance compared with Ours trained with NLL model, and much better compared with the baseline outputting equal uncertainty of 1.}
\label{table:uncertain_sdfs}
\end{table*}

We quantitatively use two metrics, Energy Score (ES) and Negative Log Likelihood (NLL) to evaluate the calibration and sharpness of the estimated SDF uncertainty in Table~\ref{table:uncertain_sdfs}. We show results for the points near the object's surface (Surface), far away from the surface (Non-Surface) by a distance threshold of 0.01, and all points (All Space) separately. 
We ablade two models trained with ES and NLL separately. Considering the lack of an uncertainty-aware shape model in the literature, we use a baseline that outputs equal uncertainty of 1. As expected, our two uncertainty models both outperform the baseline by an order of magnitude, which shows that the estimated uncertainty contains much more valid information. 
Compared with the NLL model, the ES model can output slightly better uncertainty, but an obviously better accuracy of the 3D reconstruction.
We notice the uncertainty of the points near the surface (Surface) is better calibrated than those far away from the surface (Non-Surface). The points near the surface are critical for the process of Marching Cubes to reconstruct objects' mesh and they model the uncertainty of the mesh surface. We think this is because, during the training of the Decoder, more points are sampled near the surface where the Decoder is more capable of estimating the SDFs, which benefits the uncertainty estimation too.

\subsection{Distribution of the SDF Values}

\begin{figure}[t]
\centering
\includegraphics[width=0.45\textwidth]{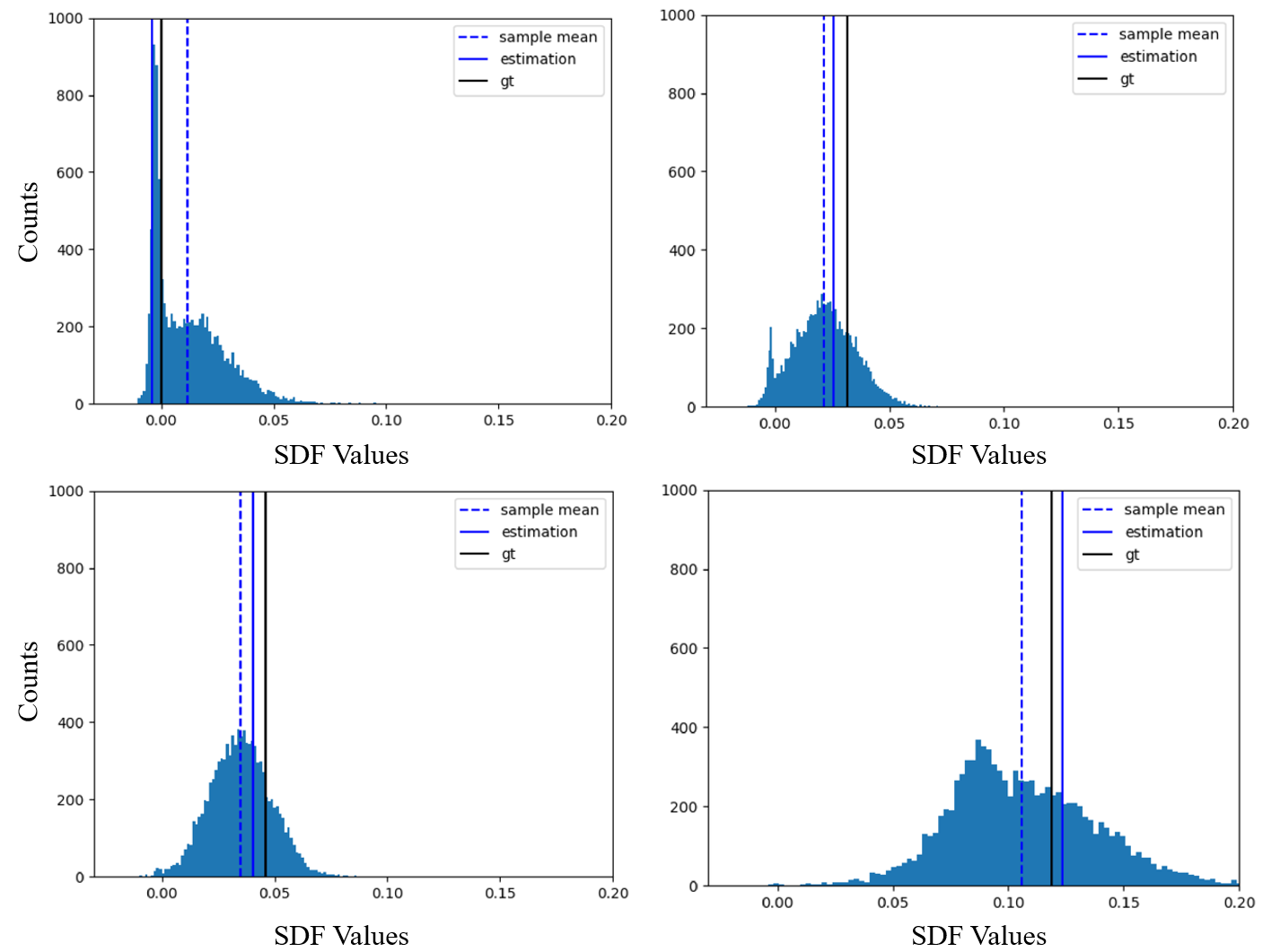}
\caption{Distribution of the SDF values after propagating the latent code distribution to the SDF space with Monte-carlo sampling. Four points with distance near and far from the surface (SDF=0) are shown. 
The groundtruth value (gt), estimated value (estimation) and the sample mean are shown. The estimation is close to the groudntruth value and the distribution covers the error.}
\label{fig:sdf-distribution}
\vspace{-1.1\baselineskip}

\end{figure}

We propagate a Gaussian distribution in the latent space through the Decoder to generate the distributions of SDF values. 
However, the Decoder, with multiple non-linear activation layers, is a highly non-linear function. Thus, the SDF distributions are not guaranteed to be Gaussian again. 
We show the distribution of the SDF values of 4 points with different distances to the object surface in Figure \ref{fig:sdf-distribution}. Their distributions are calculated through Monte-carlo sampling with a sample num of 1000. We can see that for the points far away from the surface (the two figures in the bottom), the distributions of the SDFs value only have one mode and can approximately form a Gaussian distribution. 
We notice an interesting findings for the SDF values near the surface (the two figures in the top). They do not follow a Gaussian distribution perfectly but have a peak much larger than other areas in the negative areas near the surface. 
We assume this is a bias introduced during training where the Decoder is trained with unbalanced samples of points inside and outside surface. The origin DeepSDF paper~\cite{park2019deepsdf} truncates SDF values inside the surface to $-0.01$. The trained Decoder will tend to output a narrow range of negative values, while a much larger range output space for positive values.

There are several ways to model the distribution of the SDF values depending on the downstream tasks. e.g., in the qualitative evaluation, we use the sample variance of each vertices of the mesh to approximately colorize it with uncertainty. For other tasks that need to accurately propagate uncertainty, e.g., differentiable rendering, a parametric approximation of the SDF distibution, e.g., Gaussian, will make the process easier to tackle.
It will be interesting future work to explore more expressive distribution for SDF values, e.g., Gaussian-mixture models~\cite{reynolds2009gaussian}.

\section{More Qualitative Results}

\begin{figure*}[t]
\centering
\includegraphics[width=0.9885\textwidth]{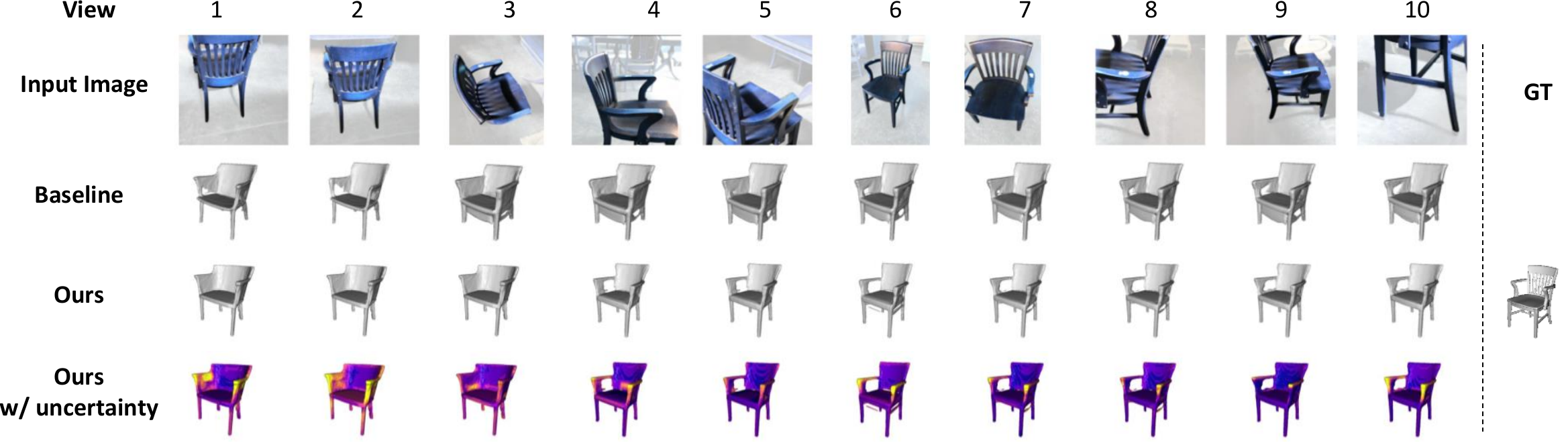}

\includegraphics[width=1.0\textwidth]{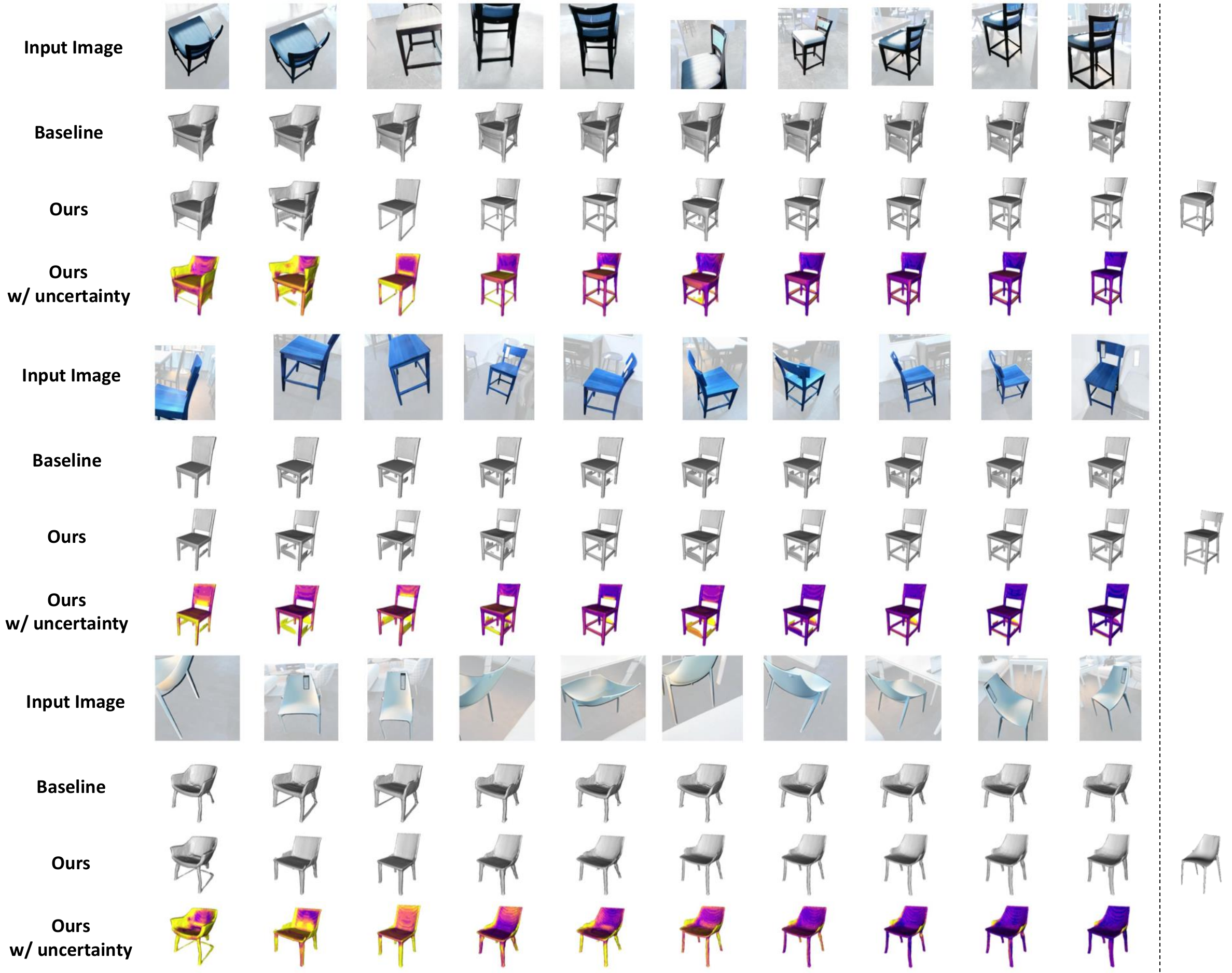}

\caption{Qualitative results of multi-view fusion on the Pix3D dataset (part 2). Results after fusing 1 to 10 input images are given. }
\label{fig:vis-fusion-part2}
\end{figure*}

We show more results of multi-view fusion in Figure~\ref{fig:vis-fusion-part2}. 
When fusing views with more information about the objects, our method decreases the uncertainty of the reconstructed shape and gets better results than the baseline. 
We notice interesting findings that some semantic parts of the objects, e.g., arms, legs, backs, and thin structures, can be well identified by uncertainty.
We also notice that for the first example, half of the 10th view including the chair arm is occluded and is very challenging, so fusing this view slightly increases the uncertainty of the arm.
It will be valuable future work to deeply investigate the relationship among the semantic parts of 3D shapes, the occluded image areas, and the uncertainty in the encoded shape latent space.

\section{Computation Analysis}

We show the inference time with a resolution of 128 for the decoder and a number of $N=10$ for the Monte Carlo sampling on an A100 GPU in Table~\ref{table:computation}. To generate the uncertainty for a mesh, only a small computation overhead is needed for the mesh vertices (2K points). However, if necessary, the uncertainty for a whole SDF field is more expensive, which consists of $128^3$ (2M) points. We inherit the design of DeepSDF~\cite{park2019deepsdf} as Decoder, which is the bottleneck of our computation. 
Depending on the downstream tasks, we can use multiple parameters as trade-off between effectiveness and efficiency, such as the sampling times and shape resolution.
Although the Monte-Carlo sampling approach used to propagate uncertainty through the decoder to the SDF grid is computationally expensive, we prove it is possible to estimate uncertainty with the decoder fixed, and leave it as future work further improvements in efficiency for global SDF uncertainty estimation., Methods such as local linearization of the nonlinear decoder, or making the decoder to direct output uncertainty can be used in this context to reduce this computational cost. 
For multi-view fusion, thanks to the fusion in the latent space, we only need 0.1 ms to fuse 10 views and get the final fused latent code with uncertainty.

\begin{table}[]
\centering
\resizebox{0.9\columnwidth}{!}{%

\begin{tabular}{@{}c|cccc@{}}
\toprule
\textbf{Modules} & \begin{tabular}[c]{@{}c@{}}SDF\\ Field\end{tabular} & Mesh & \begin{tabular}[c]{@{}c@{}}Mesh \\ w/ Uncer.\end{tabular} & \begin{tabular}[c]{@{}c@{}}SDF Field \\ w/ Uncer.\end{tabular} \\ \midrule
Encoder          & \multicolumn{4}{c}{0.06}                                                                                                                                                                \\ \midrule
Decoder          & \multicolumn{4}{c}{0.49}                                                                                                                                                                \\ \midrule
Marching Cubes   & -                                                   & 0.13 & 0.13                                                      & -                                                              \\ \midrule
Decode Uncer.    & -                                                   & -    & 0.001 x 10                                                & -                                                              \\ \midrule
Decode Uncer.    & -                                                   & -    & -                                                         & 0.49 x 10                                                      \\ \midrule
Total            & 0.55                                                & 0.68 & 0.69                                                      & 5.45                                                           \\ \bottomrule
\end{tabular}

} % resize box

\caption{Computation analysis (Unit: s). We sample $N=10$ times during Monte-carlo sampling. When generating mesh with uncertainty, only the mesh vertices (around 2K) are propagated. When generating SDF field with uncertainty, $128^3$ (2M) points are propagated. }
\label{table:computation}
\vspace{-1.1\baselineskip}
\end{table}

\section{Discussion and Future Work}

\textbf{Multi-categories}. 
In this work, we train each category (chairs, and tables) independently and separately, as the same as the previous work (DeepSDF~\cite{park2019deepsdf}, FroDO~\cite{runz2020frodo}). With the success of large language models~\cite{nichol2022point, jun2023shap} to show the effectiveness of the scaling up, and the availability of new 3D object datasets~\cite{deitke2023objaverse} much larger than ShapeNet, it will be interesting to explore an uncertainty-aware general object model that can be used for multiple categories of objects.

\textbf{Decoupling the shape decoder and the image encoder}. We take the training strategy of first training the decoder, and then fixing the decoder to train the encoder, instead of end-to-end training. 
In this way, the decoder acts as an independent shape representation in 3D that has no relationship with input modalities.   Thus, we can easily support new conditional modalities by training a new encoder, and leverage the prior knowledge stored in the latent space of the shape decoder. 
It remains valuable future work to extend to other modalities, e.g., texts, depth maps, pointclouds, by training a new corresponding encoder and fusing the results in the latent space in a similar way to the multi-view fusion. 

% {\small
% \bibliographystyle{ieee_fullname}
% \bibliography{egbib}
% }

\end{document}